\documentclass{article}

\usepackage{arxiv}

\usepackage{microtype}
\usepackage{graphicx}
\usepackage{booktabs}
\usepackage[
    colorlinks=true,
    linkcolor=blue,
    urlcolor=blue,
    citecolor=blue
]{hyperref}

\usepackage{amsfonts}
\usepackage{amsmath}
\usepackage{anyfontsize}
\usepackage{subcaption}

\DeclareMathOperator*{\argmin}{arg\,min}
\DeclareMathOperator{\Exp}{\mathbb{E}}
\DeclareMathOperator{\Var}{\mathbb{V}}
\newcommand{\norm}[1]{\left\lVert #1 \right\rVert}
\newcommand{\E}[2][]{\Exp_{#1}\!\left[ #2 \right]}
\newcommand{\Tr}[1]{\text{Tr}\left( #1 \right)}
\newcommand{\V}[2][]{\Var_{#1}\!\left[ #2 \right]}
\newcommand{\s}{\sigma}
\newcommand{\Si}{\Sigma}
\newcommand{\calL}{\mathcal{L}}

\icmltitlerunning{Deep Learning with Importance Sampling}

\begin{document}

\twocolumn[
\icmltitle{Not All Samples Are Created Equal: \\ Deep Learning with Importance Sampling}

\begin{icmlauthorlist}
\icmlauthor{Angelos Katharopoulos}{idiap,epfl}
\icmlauthor{Fran\c{c}ois Fleuret}{idiap,epfl}
\end{icmlauthorlist}

\icmlaffiliation{idiap}{Idiap Research Institute, Switzerland}
\icmlaffiliation{epfl}{\'Ecole Polytechique F\'ed\'erale de Lausanne, Switzerland}

\icmlcorrespondingauthor{Angelos Katharopoulos}{firstname.lastname@idiap.ch}

\icmlkeywords{variance reduction, importance sampling, deep learning}

\vskip 0.3in
]

\printAffiliationsAndNotice{}

\begin{abstract}
Deep neural network training spends most of the computation on
examples that are properly handled, and could be ignored.

We propose to mitigate this phenomenon with a principled importance
sampling scheme that focuses computation on ``informative'' examples,
and reduces the variance of the stochastic gradients during
training. Our contribution is twofold: first, we derive a tractable
upper bound to the per-sample gradient norm, and second we derive an
estimator of the variance reduction achieved with importance sampling,
which enables us to switch it on when it will result in an actual
speedup.

The resulting scheme can be used by changing a few lines of code in a standard
SGD procedure, and we demonstrate experimentally, on image classification, CNN
fine-tuning, and RNN training, that for a fixed wall-clock time budget, it
provides a reduction of the train losses of up to an order of magnitude and a
relative improvement of test errors between 5\% and 17\%.
\end{abstract}

\section{Introduction}

The dramatic increase in available training data has made the use of deep
neural networks feasible, which in turn has significantly improved the
state-of-the-art in many fields, in particular computer vision and natural
language processing. However, due to the complexity of the resulting
optimization problem, computational cost is now the core issue in training
these large architectures.

When training such models, it appears to any practitioner that not all samples
are equally important; many of them are properly handled after a few epochs of
training, and most could be ignored at that point without impacting the final
model. To this end, we propose a novel importance sampling scheme that
accelerates the training of any neural network architecture by focusing the
computation on the samples that will introduce the biggest change in the
parameters which reduces the variance of the gradient estimates.

For convex optimization problems, many works \cite{bordes2005fast,
zhao2015stochastic, needell2014stochastic, canevet-et-al-2016,
richtarik2013optimal} have taken advantage of the difference in importance
among the samples to improve the convergence speed of stochastic optimization
methods. On the other hand, for deep neural networks, sample
selection methods were mainly employed to generate hard negative samples for
embedding learning problems or to tackle the class imbalance problem
\cite{schroff2015facenet, Wu_2017_ICCV, simo2015discriminative}.

Recently, researchers have shifted their focus on using importance sampling to
improve and accelerate the training of neural networks \cite{alain2015variance,
loshchilov2015online, schaul2015prioritized}.  Those works, employ either the
gradient norm or the loss to compute each sample's importance. However, the
former is prohibitively expensive to compute and the latter is not a
particularly good approximation of the gradient norm.

Compared to the aforementioned works, we derive an upper bound to the per
sample gradient norm that can be computed in a single forward pass. This
results in reduced computational requirements of more than an order of
magnitude compared to \citet{alain2015variance}. Furthermore, we quantify the
variance reduction achieved with the proposed importance sampling scheme and
associate it with the batch size increment required to achieve an equivalent
variance reduction. The benefits of this are twofold, firstly we provide an
intuitive metric to predict how useful importance sampling is going to be, thus
we are able to decide when to switch on importance sampling during training. Secondly, we also
provide theoretical guarantees for speedup, when variance reduction is above a
threshold. Based on our analysis, we propose a simple to use algorithm that can
be used to accelerate the training of any neural network architecture.

Our implementation is generic and can be employed by adding a single line of code in
a standard Keras model training. We validate it on three independent tasks:
image classification,
fine-tuning and sequence classification with recurrent neural networks.
Compared to existing batch selection schemes, we show that our method
consistently achieves lower training loss and test error for equalized
wall-clock time.

\section{Related Work}

Existing importance sampling methods can be roughly categorized in methods
applied to convex problems and methods designed for deep neural networks.

\subsection{Importance Sampling for Convex Problems}

Importance sampling for convex optimization problems has been extensively
studied over the last years. \citet{bordes2005fast} developed LASVM, which is
an online algorithm that uses importance sampling to train kernelized support
vector machines.  Later, \citet{richtarik2013optimal} proposed a generalized
coordinate descent algorithm that samples coordinate sets in a way that
optimizes the algorithm's convergence rate.

More recent works \cite{zhao2015stochastic, needell2014stochastic} make a
clear connection with the variance of the gradient estimates of stochastic
gradient descent and show that the optimal sampling distribution is
proportional to the per sample gradient norm. Due to the relatively simple
optimization problems that they deal with, the authors resort to sampling
proportionally to the norm of the inputs, which in simple linear
classification is proportional to the Lipschitz constant of the per sample loss function.

Such simple importance measures do not exist for Deep Learning and the direct
application of the aforementioned theory \cite{alain2015variance}, requires
clusters of GPU workers just to compute the sampling distribution.

\subsection{Importance Sampling for Deep Learning}

Importance sampling has been used in Deep Learning mainly in the form of
manually tuned sampling schemes. \citet{bengio2009curriculum} manually design a
sampling scheme inspired by the perceived way that human children learn; in
practice they provide the network with examples of increasing difficulty in an
arbitrary manner. Diametrically opposite, it is common for deep embedding
learning to sample hard examples because of the plethora of easy non
informative ones \cite{simo2015discriminative, schroff2015facenet}.

More closely related to our work, \citet{schaul2015prioritized} and
\citet{loshchilov2015online} use the loss to create the sampling distribution.
Both approaches keep a history of losses for previously seen samples, and sample
either proportionally to the loss or based on the loss ranking. One of the main limitations of history based sampling,
is the need for tuning a large number of hyperparameters that control the
effects of ``stale'' importance scores; i.e. since the model is constantly
updated, the importance of samples fluctuate and previous observations may
poorly reflect the current situation. In particular,
\citet{schaul2015prioritized} use various forms of
smoothing for the losses and the importance sampling weights, while
\citet{loshchilov2015online} introduce a large number of hyperparameters that
control when the losses are computed, when they are sorted as well as how the
sampling distribution is computed based on the rank.

In comparison to all the above methods, our importance sampling scheme based on
an upper bound to the gradient norm has a solid theoretical basis with clear
objectives, very easy to choose hyperparameters, theoretically guaranteed
speedup and can be applied to any type of network and loss function.

\subsection{Other Sample Selection Methods}

For completeness, we mention the work of \citet{Wu_2017_ICCV}, who design a
distribution (suitable only for the distance based losses) that maximizes the
diversity of the losses in a single batch. In addition, \citet{fan2017learning}
use reinforcement learning to train a neural network that selects samples for
another neural network in order to optimize the convergence speed. Although
their preliminary results are promising, the overhead of training two networks
makes the wall-clock speedup unlikely and their proposal not as appealing.

\subsection{Stochastic Variance Reduced Gradient}

Finally, a class of algorithms that aim to accelerate the convergence of
Stochastic Gradient Descent (SGD) through variance reduction are SVRG type
algorithms \cite{johnson2013accelerating, defazio2014saga, allen2017katyusha,
lei2017non}. Although asymptotically better, those algorithms typically perform
worse than plain SGD with momentum for the low accuracy optimization setting of
Deep Learning. Contrary to the aforementioned algorithms, our proposed
importance sampling does not improve the asymptotic convergence of SGD but
results in pragmatic improvements in all the metrics given a fixed time budget.

\section{Variance Reduction for Deep Neural Networks}

Importance sampling aims at increasing the convergence speed of SGD by focusing
computation on samples that actually induce a change in the model parameters.
This formally translates into a reduced variance of the gradient estimates for
a fixed computational cost. In the following sections, we analyze how this
works and present an efficient algorithm that can be used to train any Deep
Learning model.

\subsection{Introduction to Importance Sampling}

Let $x_i$, $y_i$ be the $i$-th input-output pair from the training set,
$\Psi(\cdot; \theta)$ be a Deep Learning model parameterized by the vector
$\theta$, and $\calL(\cdot, \cdot)$ be the loss function to be minimized during
training.
The goal of training is to find
\begin{equation}
\theta^* = \argmin_\theta \frac{1}{N} \sum_{i=1}^N \calL(\Psi(x_i; \theta), y_i)
\end{equation}
where $N$ corresponds to the number of examples in the training set.

We use an SGD procedure with learning rate $\eta$, where the update at iteration $t$ depends on the sampling distribution $p^t_1, \dots, p^t_N$ and re-scaling coefficients $w^t_1, \dots, w^t_N$. Let $I_t$ be the data point sampled at that step, we have $P(I_t  = i) = p^t_i$ and
\begin{equation}
\theta_{t+1} = \theta_t - \eta w_{I_t} \nabla_{\theta_t} \calL(\Psi(x_{I_t}; \theta_t), y_{I_t})
\end{equation}

Plain SGD with uniform sampling is achieved with $w^t_i = 1$ and $p^t_i =
\frac{1}{N}$ for all $t$ and $i$.

If we define the convergence speed $S$ of SGD as the reduction of the distance
of the parameter vector $\theta$ from the optimal parameter vector $\theta^*$
in two consecutive iterations $t$ and $t+1$
\begin{equation} \label{eq:convergence_speed_first_norms}
S = -\E[P_t]{\norm{\theta_{t+1} - \theta^*}_2^2 - \norm{\theta_{t} - \theta^*}_2^2},
\end{equation}
and if we have $w_i = \frac{1}{N p_i}$ such that
\begin{eqnarray}
\lefteqn{\E[P_t]{w_{I_t} \nabla_{\theta_t} \calL(\Psi(x_{I_t}; \theta_t), y_{I_t})}} \\
    & \quad = \nabla_{\theta_t} \frac{1}{N}\sum_{i=1}^N \calL(\Psi(x_i; \theta_t), y_i),
\end{eqnarray}
and set $G_i = w_i \nabla_{\theta_t}\calL(\Psi(x_i; \theta_t), y_i)$,
then we get (this is a different derivation of the result by \citealp{wang2016accelerating})
\begin{equation} \label{eq:convergence_speed_first} 
\begin{aligned}
S
    & = -\E[P_t]{
        \left(\theta_{t+1} \!-\! \theta^*\right)^T\! \left(\theta_{t+1} \!-\! \theta^*\right) -
        \left(\theta_t \!-\! \theta^*\right)^T\! \left(\theta_t \!-\! \theta^*\right)
    } \\
    & = -\E[P_t]{
        \theta_{t+1}^T\! \theta_{t+1} \!-\! 2 \theta_{t+1} \theta^* -
        \theta_{t}^T\! \theta_{t} + 2 \theta_{t} \theta^*
    } \\
    & = -\E[P_t]{
        \left(\theta_t \!-\! \eta G_{I_t} \right)^T\! \left(\theta_t \!-\! \eta G_{I_t} \right) +
        2 \eta G_{I_t}^T\! \theta^* \!-\! \theta_t^T\! \theta_t
    } \\
    & = -\E[P_t]{
        -2 \eta \left(\theta_t \!-\! \theta^*\right) G_{I_t} + \eta^2 G_{I_t}^T\! G_{I_t}
    } \\
    & = 2 \eta \left(\theta_t \!-\! \theta^*\right) \E[P_t]{G_{I_t}} -
        \eta^2 \E[P_t]{G_{I_t}}^T\! \E[P_t]{G_{I_t}} - \\
    & \quad \, \eta^2 \Tr{\V[P_t]{G_{I_t}}}
\end{aligned}
\end{equation}

Since the first two terms, in the last expression, are the speed of batch
gradient descent, we observe that it is possible to gain a speedup by sampling
from the distribution that minimizes $\Tr{\V[P_t]{G_{I_t}}}$. Several works
\cite{needell2014stochastic, zhao2015stochastic, alain2015variance} have shown
the optimal distribution to be proportional to the per-sample gradient norm.
However, computing this distribution is computationally prohibitive.

\subsection{Beyond the Full Gradient Norm}

Given an upper bound $\hat{G}_i \geq \norm{\nabla_{\theta_t}\calL(\Psi(x_i;
\theta_t), y_i)}_2$ and due to 
\begin{align}
\argmin_{P} \Tr{\V[P_t]{G_{I_t}}} = \argmin_{P} \E[P_t]{\norm{G_{I_t}}_2^2},
\end{align}
we propose to relax the optimization problem in the following way
\begin{align}
\min_{P} \E[P_t]{\norm{G_{I_t}}_2^2} \leq
    \min_{P} \E[P_t]{w_{I_t}^2 \hat{G}_{I_t}^2}. \label{eq:relaxation}
\end{align}
The minimizer of the second term of equation \ref{eq:relaxation},
similar to the first term, is $p_i \propto \hat{G}_i$. All that remains, is to
find a proper expression for $\hat{G}_i$ which is significantly easier to
compute than the norm of the gradient for each sample.

In order to continue with the derivation of our upper bound $\hat{G}_i$, let us introduce some notation specific to a multi-layer perceptron. Let $\theta^{(l)} \in \mathbb{R}^{M_l \times M_{l-1}}$ be the weight matrix for layer $l$ and $\s^{(l)}(\cdot)$ be a Lipschitz continuous activation function. Then, let
\begin{align}
    x^{(0)}              & = x                                                                                                        \\
    z^{(l)}              & = \theta^{(l)} \, x^{(l-1)}                                                                                \\
    x^{(l)}              & = \s^{(l)}(z^{(l)})                                                                                        \\
    \Psi(x; \Theta)      & = x^{(L)}
\end{align}
Although our notation describes
simple fully connected neural networks without bias, our analysis holds for any
affine operation followed by a slope-bounded non-linearity ($|\s'(x)| \leq K$).
With
\begin{align}
\Si_l'(z)                & = diag\left(\sigma'^{(l)}(z_1), \dots, \sigma'^{(l)}(z_{M_l})\right),                      \\
\Delta_i^{(l)}             & = \Si_l'(z_i^{(l)}) \theta_{l+1}^T \dots \Si_{L-1}'(z_i^{(L-1)}) \theta_L^T,                                         \\
\nabla_{x_i^{(L)}} \calL & = \nabla_{x_i^{(L)}} \calL(\Psi(x_i; \Theta), y_i)
\end{align}
we get
\begin{eqnarray}
\lefteqn{\norm{\nabla_{\theta_l} \calL(\Psi(x_i; \Theta), y_i)}_2}                                                                    \\
                         &  =    & \norm{\left( \Delta_i^{(l)} \Si_L'(z_i^{(L)}) \nabla_{x_i^{(L)}}\calL \right) \left(x_i^{(l-1)}\right)^T}_2 \\
                         &  \leq & \norm{\Delta_i^{(l)}}_2 \norm{\Si_L'(z_i^{(L)}) \nabla_{x_i^{(L)}}\calL}_2 \norm{x_i^{(l-1)}}_2             \\
                         &  \leq & \underbrace{\max_{l, i}\left( \norm{x_i^{(l-1)}}_2 \norm{\Delta_i^{(l)}}_2 \right)}_{\rho} \norm{\Si_L'(z_i^{(L)}) \nabla_{x_i^{(L)}}\calL}_2
                         \label{eq:tentative_bound}
\end{eqnarray}


Various weight initialization \cite{Glorot10} and activation normalization
techniques \cite{batch-normalization-icml2015,ba2016layernormalization}
uniformise the activations across samples.
As a result, the variation of the gradient norm is mostly
captured by the gradient of the loss function with respect to the
pre-activation outputs of the last layer of our neural
network. Consequently we can derive the following upper bound to the
gradient norm of all the parameters
\begin{align}
    \norm{\nabla_{\Theta} \calL(\Psi(x_i; \Theta), y_i)}_2 \leq
        \underbrace{L \rho \norm{\Si_L'(z_i^{(L)}) \nabla_{x_i^{(L)}}\calL}_2}_{\hat{G}_i} \label{eq:upper_bound},
\end{align}
which is marginally more difficult to compute than the value of the loss since it can be computed in a closed form in terms of $z^{(L)}$.
However, our upper bound depends on the time step $t$, thus we cannot generate
a distribution once and sample from it during training. This is intuitive
because the importance of each sample changes as the model changes.

\subsection{When is Variance Reduction Possible?}\label{sec:when-vari-reduct}

Computing the importance score from equation~\ref{eq:upper_bound} is more than
an order of magnitude faster compared to computing the gradient norm for each
sample.  Nevertheless, it still costs one forward pass through the network and
can be wasteful. For instance, during the first iterations of training, the
gradients with respect to every sample have approximately equal norm; thus we
would waste computational resources trying to sample from the uniform
distribution. In addition, computing the importance score for the whole dataset
is still prohibitive and would render the method unsuitable for online
learning.

In order to solve the problem of computing the importance for the whole
dataset, we \textit{pre-sample} a large batch of data points, compute the
sampling distribution for that batch and re-sample a smaller batch with
replacement. The above procedure upper bounds both the speedup and variance
reduction. Given a large batch consisting of $B$ samples and a small one consisting
of $b$, we can achieve a maximum variance reduction of $\frac{1}{b} -
\frac{1}{B}$ and a maximum speedup of $\frac{B+3b}{3B}$ assuming that the
backward pass requires twice the amount of time as the forward pass.

Due to the large cost of computing the importance per sample, we only perform
importance sampling when we know that the variance of the gradients can be
reduced. In the following equation, we show that the variance reduction is
proportional to the squared $L_2$ distance of the sampling distribution, $g$,
to the uniform distribution $u$. Due to lack of space, the complete derivation
is included in the supplementary material. Let $g_i \propto
\norm{\nabla_{\theta_t} \calL(\Psi(x_i; \theta_t), y_i)}_2 = \norm{G_i}_2$ and
$u = \frac{1}{B}$ the uniform probability.
\begin{align}
& \Tr{\V[u]{G_i}} - \Tr{\V[g]{w_i G_i}} \\
& = \E[u]{\norm{G_i}_2^2} - \E[g]{w_i^2 \norm{G_i}_2^2} \\
& = \left(\frac{1}{B} \sum_{i=1}^B \norm{G_i}_2\right)^2 B \norm{g - u}_2^2.
\label{eq:distance}
\end{align}

Equation \ref{eq:distance} already provides us with a useful metric to decide
if the variance reduction is significant enough to justify using importance
sampling. However, choosing a suitable threshold for the $L_2$ distance squared
would be tedious and unintuitive. We can do much better by dividing the
variance reduction with the original variance to derive the increase in the
batch size that would achieve an equivalent variance reduction. Assuming that
we increase the batch size by $\tau$, we achieve variance reduction
$\frac{1}{\tau}$; thus we have\footnote{In the first version we mistakenly
assume $\frac{1}{\tau^2}$ which made the algorithm unnecessarily conservative.
All the experiments are run using the square root of line 17 in Algorithm
\ref{alg:training}.}
\begin{align}
& \frac{\left(\frac{1}{B} \sum_{i=1}^B \norm{G_i}_2\right)^2 B \norm{g - u}_2^2}
    {\Tr{\V[u]{G_i}}} \geq \\
& \frac{\left(\frac{1}{B} \sum_{i=1}^B \norm{G_i}_2\right)^2 B \norm{g - u}_2^2}
    {\frac{1}{B} \sum_{i=1}^B \norm{G_i}_2^2} = \\
& \frac{1}{\sum_{i=1}^B g_i^2} \norm{g - u}_2^2 = 1 - \frac{1}{\tau} \iff \\
& \frac{1}{\tau} = 1 - \frac{1}{\sum_{i=1}^B g_i^2} \norm{g - u}_2^2
\label{eq:batch_increment}
\end{align}
Using equation \ref{eq:batch_increment}, we have a hyperparameter that
is very easy to select and can now design our training procedure which is
described in pseudocode in algorithm \ref{alg:training}. Computing $\tau$ from
equation \ref{eq:batch_increment} allows us to have guaranteed speedup when $B
+ 3b < 3 \tau b$. However, as it is shown in the experiments, we can use
$\tau_{th}$ smaller than $\frac{B+3b}{3b}$ and still get a significant speedup.

\begin{algorithm}
\caption{Deep Learning with Importance Sampling} \label{alg:training}
\begin{algorithmic}[1]
    \STATE Inputs $B, b, \tau_{th}, a_{\tau}, \theta_0$
    \STATE $t \gets 1$
    \STATE $\tau \gets 0$
    \REPEAT
        \IF{$\tau > \tau_{th}$}
            \STATE $\mathcal{U} \gets B$ uniformly sampled datapoints
            \STATE $g_i \propto \hat{G_i} \quad \forall i \in \mathcal{U}$ 
                according to eq~\ref{eq:upper_bound}
            \STATE $\mathcal{G} \gets b$ datapoints sampled with $g_i$
                from $\mathcal{U}$
            \STATE $w_i \gets \frac{1}{B g_i} \quad \forall i
                \in \mathcal{G}$
            \STATE $\theta_t \gets \text{sgd\_step}(w_i,
                \mathcal{G}, \theta_{t-1})$
        \ELSE
            \STATE $\mathcal{U} \gets b$ uniformly sampled datapoints
            \STATE $w_i \gets 1 \quad \forall i \in \mathcal{U}$
            \STATE $\theta_t \gets \text{sgd\_step}(w_i,
                \mathcal{U}, \theta_{t-1})$
            \STATE $g_i \propto \hat{G_i} \quad \forall i \in \mathcal{U}$
                \label{alg:line:uniform_gi}
        \ENDIF
        \STATE $\tau \gets a_{\tau} \tau + (1-a_{\tau}) \left(1 - \frac{1}{\sum_{i} g_i^2} \norm{g - \frac{1}{|\mathcal{U}|}}_2^2\right)^{-1}$
    \UNTIL{convergence}
\end{algorithmic}
\end{algorithm}

The inputs to the algorithm are the pre-sampling size $B$, the batch size $b$,
the equivalent batch size increment after which we start importance sampling
$\tau_{th}$ and the exponential moving average parameter $a_{\tau}$ used to
compute a smooth estimate of $\tau$. $\theta_0$ denotes the initial parameters
of our deep network. We would like to point out that in line
\comment{\ref{alg:line:uniform_gi}} 15 of the algorithm, we compute $g_i$ for
free since we have done the forward pass in the previous step.

The only parameter that has to be explicitly defined for our algorithm is the
pre-sampling size $B$ because $\tau_{th}$ can be set using
equation~\ref{eq:batch_increment}. We provide a small ablation study for $B$ in
the supplementary material.

\section{Experiments}

In this section, we analyse experimentally the performance of the proposed
importance sampling scheme based on our \emph{upper-bound} of the gradient
norm. In the first subsection, we compare the variance reduction achieved with
our upper bound to the theoretically maximum achieved with the true gradient
norm. We also compare against sampling based on the loss, which is commonly
used in practice. Subsequently, we conduct experiments which demonstrate that
we are able to achieve non-negligible wall-clock speedup for a variety of tasks
using our importance sampling scheme.

In all the subsequent sections, we use \emph{uniform} to refer to the usual
training algorithm that samples points from a uniform distribution, we use
\emph{loss} to refer to algorithm \ref{alg:training} but instead of sampling
from a distribution proportional to our upper-bound to the gradient norm
$\hat{G}_i$ (equations \ref{eq:relaxation} and \ref{eq:upper_bound}), we sample
from a distribution proportional to the loss value and finally
\emph{upper-bound} to refer to our proposed method. All the other baselines
from published methods are referred to using the names of the authors.

In addition to batch selection methods, we compare with various SVRG
implementations including the accelerated Katyusha \cite{allen2017katyusha} and
the online SCSG \cite{lei2017non} method. In all cases, SGD with uniform
sampling performs significantly better. Due to lack of space, we report the
detailed results in the supplementary material.

Experiments were conducted using Keras~\cite{chollet2015} with TensorFlow
\cite{abadi2016tensorflow}, and the code can be found at
\url{http://github.com/idiap/importance-sampling}. For all
the experiments, we use Nvidia K80 GPUs and the reported time is calculated by
subtracting the timestamps before starting one epoch and after finishing one;
thus it includes the time needed to transfer data between CPU and GPU memory.


Our implementation provides a wrapper around models that substitutes
the standard uniform sampling with our importance-sampling
method. This means that {\it adding a single line of code} to call
this wrapper before actually fitting the model is sufficient to switch
from the standard uniform sampling to our importance-sampling
scheme. And, as specified in \S~\ref{sec:when-vari-reduct} and
Algorithm~\ref{alg:training}, our procedure reliably estimates at
every iteration if the importance sampling will provide a speed-up and
sticks to uniform sampling otherwise.


\begin{figure}[!ht]
    \includegraphics[width=\linewidth]{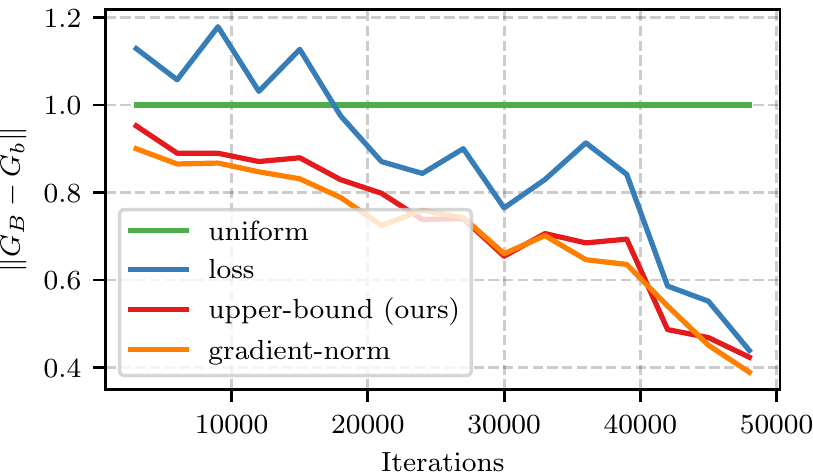}
    \caption{The y-axis denotes the $L_2$ distance of the average gradient of
    the large batch ($G_B$) and the average gradient of the small batch ($G_b$)
    normalized with the distance achieved by uniform sampling. The sampling of
    the small batch is done $10$ times and the reported results are the
    average. The details of the experimental setup are given in
    \S~\ref{sec:ablation}.}
    \label{fig:variance_reduction}
\end{figure}

\begin{figure}[!ht]
    \includegraphics[width=\linewidth]{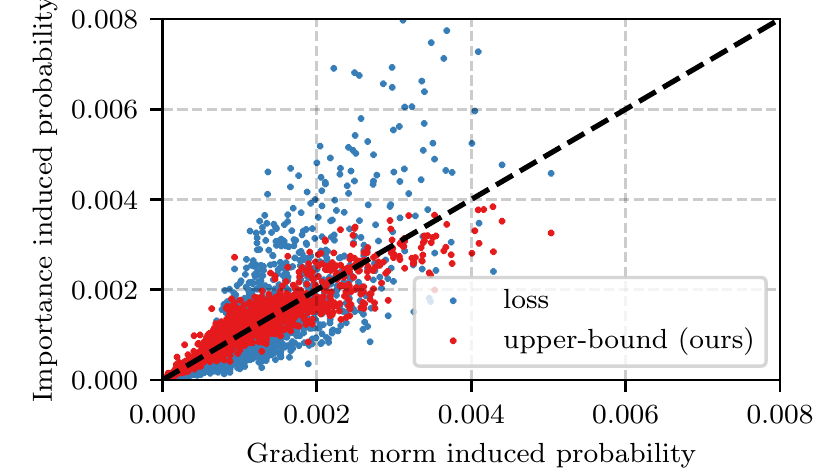}
    \caption{The probabilities generated with the \emph{loss} and our
    \emph{upper-bound} are plotted against the ideal probabilities produced by
    the \emph{gradient-norm}. The black line denotes perfect correlation. The
    details of the experimental setup are given in \S~\ref{sec:ablation}.}
    \label{fig:gnorm_fit}
    \vspace{-1em}
\end{figure}

\subsection{Ablation study} \label{sec:ablation}

As already mentioned, several works~\cite{loshchilov2015online,
schaul2015prioritized} use the loss value, directly or indirectly, to generate
sampling distributions. In this section, we present experiments that validate
the superiority of our method with respect to the loss in terms of variance
reduction. For completeness, in the supplementary material we include a
theoretical analysis that explains why sampling based on the loss also
achieves variance reduction during the late stages of training.

Our experimental setup is as follows: we train a wide residual network
\cite{zagoruyko2016wrn} on the CIFAR100 dataset~\cite{krizhevsky2009learning},
following closely the training procedure of~\citet{zagoruyko2016wrn} (the
details are presented in \S~\ref{sec:cifar}). Subsequently, we sample $1,024$
images uniformly at random from the dataset. Using the weights of the trained
network, at intervals of $3,000$ updates, we resample $128$ images from the
large batch of $1,024$ images using \emph{uniform} sampling or importance
sampling with probabilities proportional to the \emph{loss}, our
\emph{upper-bound} or the \emph{gradient-norm}. The \emph{gradient-norm} is
computed by running the backpropagation algorithm with a batch size of 1.

Figure~\ref{fig:variance_reduction} depicts the variance reduction achieved
with every sampling scheme in comparison to \emph{uniform}. We measure this
directly as the distance between the mini-batch gradient and the batch gradient
of the $1,024$ samples. For robustness we perform the sampling $10$ times and
report the average. We observe that our upper bound and the gradient norm
result in very similar variance reduction, meaning that the bound is relatively
tight and that the produced probability distributions are highly correlated.
This can also be deduced
by observing figure~\ref{fig:gnorm_fit}, where the probabilities proportional
to the \emph{loss} and the \emph{upper-bound} are plotted against the optimal
ones (proportional to the \emph{gradient-norm}). We observe that our upper
bound is almost perfectly correlated with the gradient norm, in stark contrast
to the loss which is only correlated at the regime of very small gradients.
Quantitatively the sum of squared error of $16,384$ points in
figure~\ref{fig:gnorm_fit} is $0.017$ for the \emph{loss} and $0.002$ for our
proposed upper bound.

Furthermore, we observe that sampling hard examples (with high loss), increases
the variance, especially in the beginning of training. Similar behaviour has
been observed in problems such as embedding learning where semi-hard sample
mining is preferred over sampling using the loss~\cite{Wu_2017_ICCV,
schroff2015facenet}.

\subsection{Image classification} \label{sec:cifar}

\begin{figure*}[!ht]
    \centering
    \begin{subfigure}[t]{0.49\textwidth}
        \includegraphics[width=\textwidth]{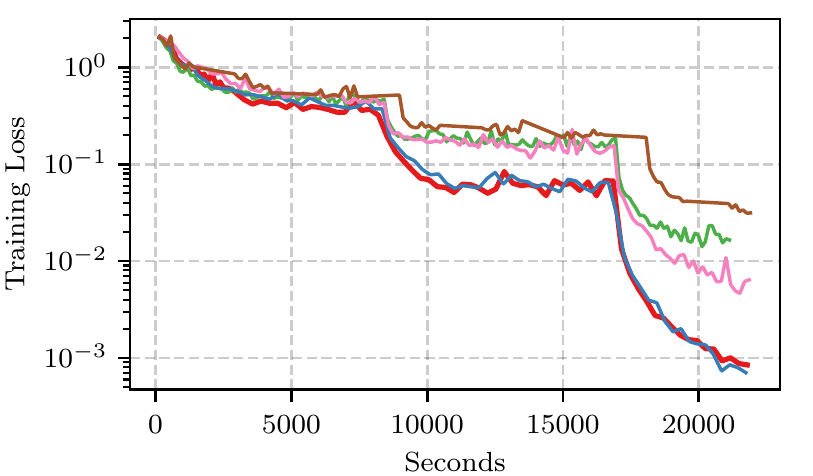}
        \caption{CIFAR10 Training Loss}
    \end{subfigure}
    ~
    \begin{subfigure}[t]{0.49\textwidth}
        \includegraphics[width=\textwidth]{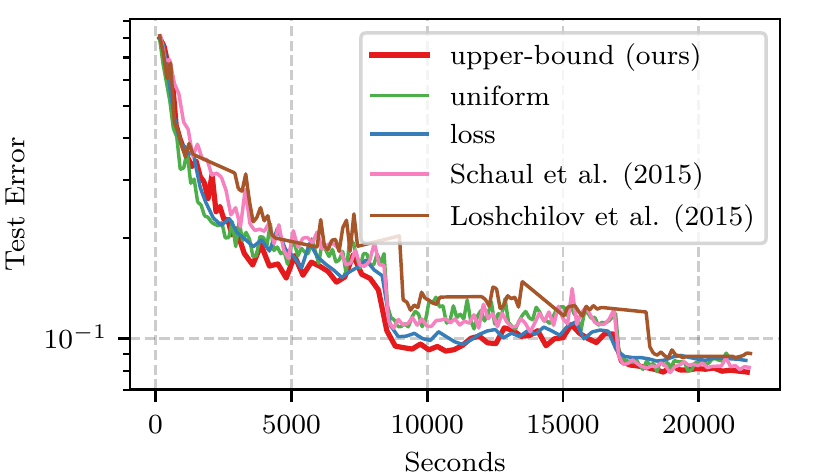}
        \caption{CIFAR10 Test Error}
    \end{subfigure} \\
    \begin{subfigure}[t]{0.49\textwidth}
        \includegraphics[width=\textwidth]{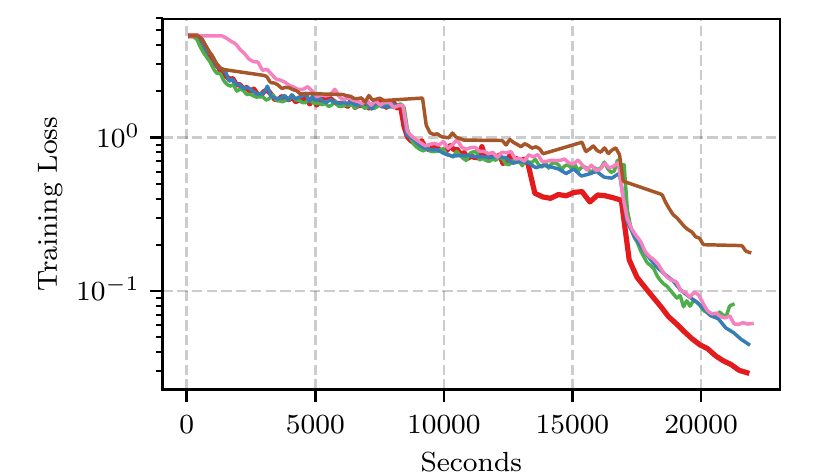}
        \caption{CIFAR100 Training Loss}
    \end{subfigure}
    ~
    \begin{subfigure}[t]{0.49\textwidth}
        \includegraphics[width=\textwidth]{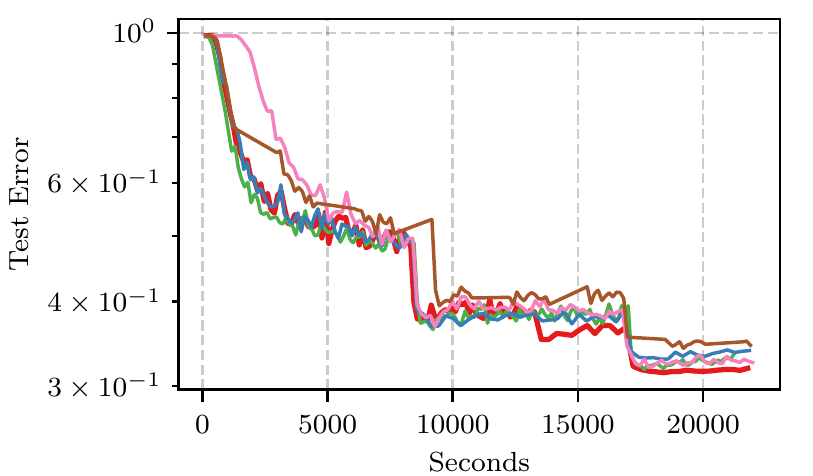}
        \caption{CIFAR100 Test Error}
    \end{subfigure}
    \caption{Comparison of importance sampling using the
    \emph{upper-bound} with \emph{uniform} and loss based importance
    sampling. The details of the training procedure are given in
    \S~\ref{sec:cifar}. Our proposed scheme is the only one achieving a
    speedup on CIFAR100 and results in 5\% smaller test error. All
    presented results are averaged across $3$ independent runs.}
    \label{fig:cifar}
\end{figure*}

In this section, we use importance sampling to train a
residual network on CIFAR10 and CIFAR100. We follow the experimental setup of
\citet{zagoruyko2016wrn}, specifically we train a wide resnet 28-2 with SGD
with momentum. We use batch size $128$, weight decay $0.0005$, momentum $0.9$,
initial learning rate $0.1$ divided by $5$ after $20,000$ and $40,000$
parameter updates. Finally, we train for a total of $50,000$ iterations. In
order for our history based baselines to be compatible with the data
augmentation of the CIFAR images, we pre-augment both datasets to generate $1.5
\times 10^6$ images for each one. Our method
does not have this limitation since it can work on infinite datasets in a true
online fashion. To compare between methods, we use a learning rate schedule
based on wall-clock time and we also fix the total seconds available for
training. A faster method should have smaller training loss and test error
given a specific time during training.

For this experiment, we compare the proposed method to \emph{uniform},
\emph{loss}, online batch selection by~\citet{loshchilov2015online} and the
history based sampling of~\citet{schaul2015prioritized}. For the method of
\citet{schaul2015prioritized}, we use their proportional sampling since the
rank based is very similar to~\citet{loshchilov2015online} and we select the
best parameters from the grid $a = \{0.1, 0.5, 1.0\}$ and $\beta = \{0.5,
1.0\}$. Similarly, for online batch selection, we use $s = \{1, 10, 10^2\}$ and
a recomputation of all the losses every $r = \{600, 1200, 3600\}$ updates.

For our method, we use a presampling size of $640$. One of the goals of this
experiment is to show that even a smaller reduction in variance can effectively
stabilize training and provide wall-clock time speedup; thus we set $\tau_{th}
= 1.5$. We perform $3$ independent runs and report the average.

The results are depicted in figure~\ref{fig:cifar}. We observe that in the
relatively easy CIFAR10 dataset, all methods can provide some speedup over
uniform sampling. However, for the more complicated
CIFAR100, only sampling with our proposed \emph{upper-bound} to the
gradient norm reduces the variance of the gradients and provides faster
convergence. Examining the training evolution in detail, we observe that on
CIFAR10 our method is the only one that achieves a significant improvement in
the test error even in the first stages of training ($4,000$ to $8,000$
seconds). Quantitatively, on CIFAR10 we achieve more than an order of magnitude
lower training loss and $8\%$ lower test error from $0.087$ to $0.079$ while on
CIFAR100 approximately $3$ times lower training loss and $5\%$ lower test error
from $0.34$ to $0.32$ compared to \emph{uniform} sampling.

At this point, we would also like to discuss the performance of the \emph{loss}
compared to other methods that also select batches based on this metric. Our
experiments show, that using ``fresh'' values for the loss combined with a
warmup stage so that importance sampling is not started too early outperforms
all the other baselines on the CIFAR10 dataset.

\subsection{Fine-tuning} \label{sec:mit67}

\begin{figure*}
    \begin{subfigure}[t]{0.49\textwidth}
        \includegraphics{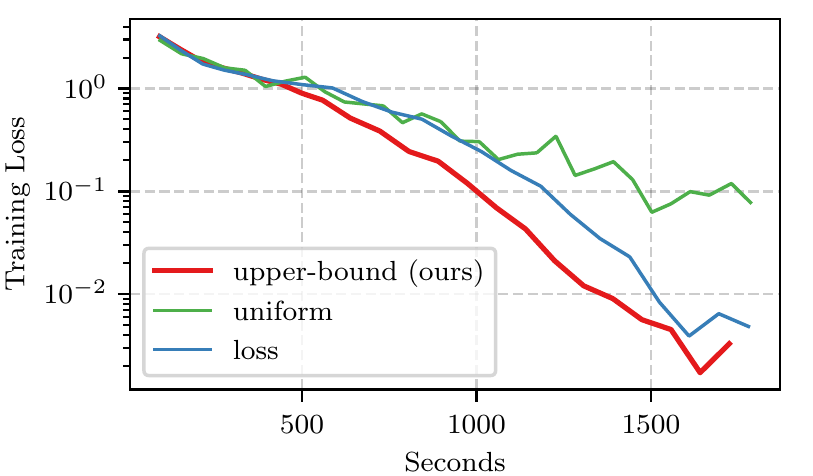}
    \end{subfigure}
    ~
    \begin{subfigure}[t]{0.49\textwidth}
        \includegraphics{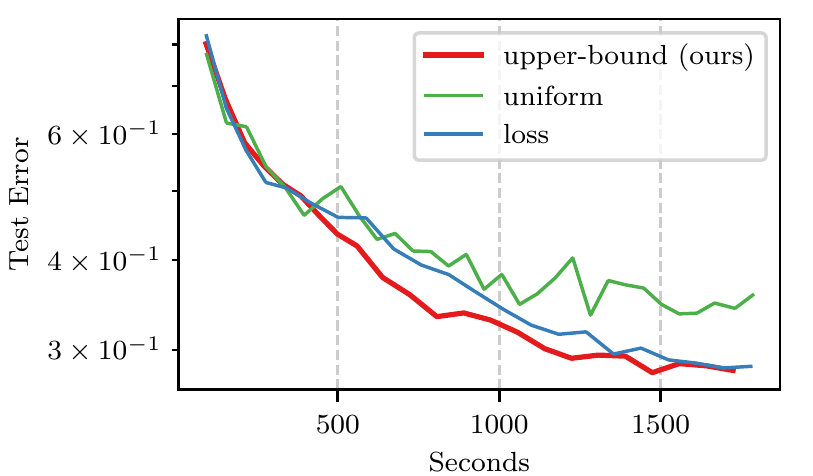}
    \end{subfigure}
    \caption{Comparison of importance sampling for fine-tuning on MIT67
    dataset. The details of the training procedure are given in
    \S~\ref{sec:mit67}. Our proposed algorithm converges very quickly to
    $28.06\%$ test error in approximately half an hour, a relative reduction of
    $17\%$ to uniform sampling. For robustness, the results are averaged across
    $3$ independent runs.}
    \label{fig:finetuning}
    \vspace{-1em}
\end{figure*}

Our second experiment shows the application of importance sampling to the
significant task of fine tuning a pre-trained large neural network on a new
dataset. This task is of particular importance because there exists an
abundance of powerful models pre-trained on large datasets such as ImageNet
\cite{imagenet_cvpr09}.

Our experimental setup is the following, we fine-tune a
ResNet-50~\cite{He2015} previously trained on ImageNet. We replace the last
classification layer and then train the whole network end-to-end to classify
indoor images among 67 possible categories~\cite{quattoni2009recognizing}. We
use SGD with learning rate $10^{-3}$ and momentum $0.9$. We set the batch size
to $16$ and for our importance sampling algorithm we pre-sample $48$. The
variance reduction threshold is set to $2$ as designated by
equation~\ref{eq:batch_increment}.

To assess the performance of both our algorithm and our gradient norm
approximation, we compare the convergence speed of our importance sampling
algorithm using our \emph{upper-bound} and using the \emph{loss}. Once again,
for robustness, we run $3$ independent runs and report the average.

The results of the experiment are depicted in figure~\ref{fig:finetuning}. As
expected, importance sampling is very useful for the task of fine-tuning since
a lot of samples are handled correctly very early in the training process. Our
\emph{upper-bound}, once again, greatly outperforms sampling proportionally to
the loss when the network is large and the problem is non trivial. Compared to
uniform sampling, in just half an hour importance sampling has converged close
to the best performance ($28.06\%$ test error) that can be expected on this
dataset without any data augmentation or multiple crops~\cite{razavian2014cnn},
while uniform achieves only $33.74\%$.

\subsection{Pixel by Pixel MNIST} \label{sec:mnist}

\begin{figure*}[!ht]
    \begin{subfigure}[t]{0.49\textwidth}
        \includegraphics{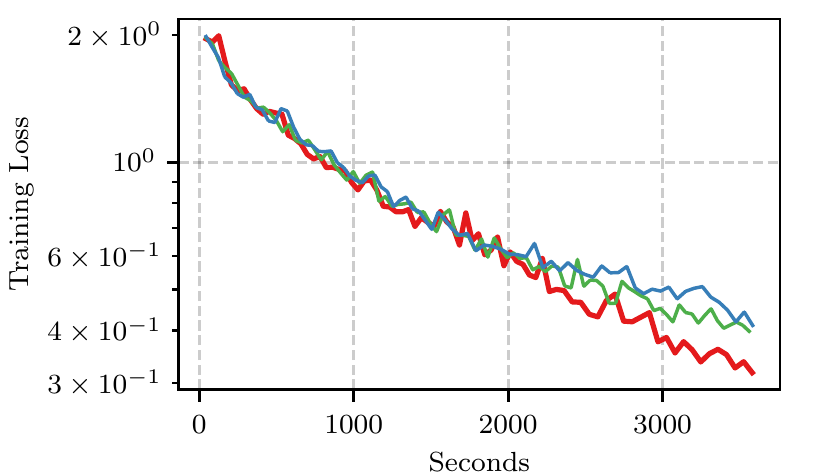}
    \end{subfigure}
    ~
    \begin{subfigure}[t]{0.49\textwidth}
        \includegraphics{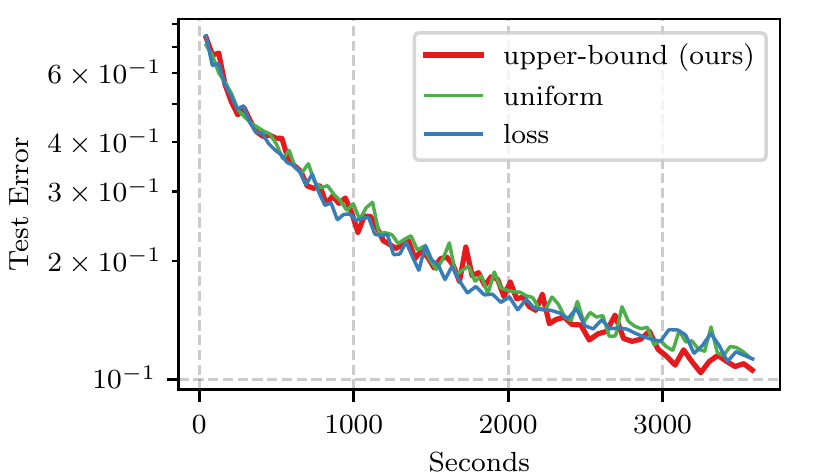}
    \end{subfigure}
    \caption{Comparison of importance sampling on pixel-by-pixel MNIST with an
    LSTM. The details of the training procedure are given in
    \S~\ref{sec:mnist}. Our proposed algorithm speeds up training and achieves
    $7\%$ lower test error in one hour of training ($0.1055$ compared to
    $0.1139$). We observe that sampling proportionally to the loss actually
    hurts convergence in this case.}
    \label{fig:pixbypix}
    \vspace{-0.5em}
\end{figure*}

To showcase the generality of our method, we use our
importance sampling algorithm to accelerate the training of an LSTM in a
sequence classification problem. We use the pixel by pixel classification of
randomly permuted MNIST digits~\cite{lecun2010mnist}, as defined by
\citet{le2015simple}. The problem may seem trivial at first, however as shown
by \citet{le2015simple} it is particularly suited to benchmarking the training
of recurrent neural networks, due to the long range dependency problems
inherent in the dataset ($784$ time steps).

For our experiment, we fix a permutation
matrix for all the pixels to generate a training set of $60,000$ samples with
$784$ time steps each. Subsequently, we train an LSTM
\cite{hochreiter1997long} with $128$ dimensions in the hidden space,
$\text{tanh}(\cdot)$ as an activation function and $\text{sigmoid}(\cdot)$ as
the recurrent activation function. Finally, we use a linear classifier on top
of the LSTM to choose a digit based on the hidden representation.
To train the aforementioned architecture, we use the Adam optimizer
\cite{kingma2014adam} with a learning rate of $10^{-3}$ and a batch size of
$32$. We have also found gradient clipping to be necessary for the training
not to diverge; thus we clip the norm of all gradients to $1$.

The results of the experiment are depicted in figure~\ref{fig:pixbypix}. Both
for the \emph{loss} and our proposed \emph{upper-bound}, importance sampling
starts at around $2,000$ seconds by setting $\tau_{th}=1.8$ and the presampling
size to $128$. We could set $\tau_{th}=2.33$
(equation~\ref{eq:batch_increment}) which would only result in our algorithm
being more conservative and starting importance sampling later. We clearly
observe that sampling proportionally to the loss hurts the convergence in this
case. On the other hand, our algorithm achieves $20\%$ lower training loss and
$7\%$ lower test error in the given time budget.

\section{Conclusions}

We have presented an efficient algorithm for accelerating the training of deep
neural networks using importance sampling. Our algorithm takes advantage of a
novel upper bound to the gradient norm of any neural network that can be
computed in a single forward pass. In addition, we show an equivalence of the
variance reduction with importance sampling to increasing the batch
size; thus we are able to quantify both the variance reduction and the speedup
and intelligently decide when to stop sampling uniformly.

Our experiments show that our algorithm is effective in reducing the training
time for several tasks both on image and sequence data. More importantly, we
show that not all data points matter equally in the duration of training, which
can be exploited to gain a speedup or better quality gradients or both.

Our analysis opens several avenues of future research. The two most important
ones that were not investigated in this work are automatically tuning the
learning rate based on the variance of the gradients and decreasing the batch
size. The variance of the gradients can be kept stable by increasing the
learning rate proportionally to the batch increment or by decreasing the number
of samples for which we compute the backward pass. Thus, we can speed up
convergence by increasing the step size or reducing the time per update.

\section{Acknowledgement}

This work is supported by the Swiss National Science Foundation under grant
number FNS-30209 ``ISUL''.

\bibliography{references}
\bibliographystyle{icml2018}

\clearpage
\twocolumn[
    \standalonetitle{Appendix}
]
\appendix
\section{Differences of variances}

In the following equations we quantify the variance reduction achieved with
importance sampling using the gradient norm. Let $g_i \propto
\norm{\nabla_{\theta_t} \calL(\Psi(x_i; \theta_t), y_i)}_2 = \norm{G_i}_2$ and
$u = \frac{1}{B}$ the uniform probability.

We want to compute
\begin{multline}
\Tr{\V[u]{G_i}} - \Tr{\V[g]{w_i G_i}} \\
= \E[u]{\norm{G_i}_2^2} - \E[g]{w_i^2 \norm{G_i}_2^2}.
\end{multline}
Using the fact that $w_i = \frac{1}{B g_i}$ we have
\begin{align}
\E[g]{w_i^2 \norm{G_i}_2^2} = \left( \frac{1}{B} \sum_{i=1}^B \norm{G_i}_2 \right)^2 ,
\end{align}
thus
\begin{align}
& \Tr{\V[u]{G_i}} - \Tr{\V[g]{w_i G_i}} \\
& = \frac{1}{B} \sum_{i=1}^B \norm{G_i}_2^2 - \left( \frac{1}{B} \sum_{i=1}^B
    \norm{G_i}_2 \right)^2 \\
& = \frac{\left(\sum_{i=1}^B \norm{G_i}_2\right)^2}{B^3} \sum_{i=1}^B \left(
    B^2 \frac{\norm{G_i}_2^2}{(\sum_{i=1}^B \norm{G_i}_2)^2} - 1 \right) \\
& = \frac{\left(\sum_{i=1}^B \norm{G_i}_2\right)^2}{B} \sum_{i=1}^B \left(
    g_i^2 - u^2 \right) \label{eq:sup_diff_var_sq}.
\end{align}
Completing the squares at equation \ref{eq:sup_diff_var_sq} and using the fact
that $\sum_{i=1}^B u = 1$ we complete the derivation.
\begin{align}
& \Tr{\V[u]{G_i}} - \Tr{\V[g]{w_i G_i}} \\
& = \frac{\left(\sum_{i=1}^B \norm{G_i}_2\right)^2}{B} \sum_{i=1}^B \left( g_i
    - u \right)^2 \\
& = \left(\frac{1}{B} \sum_{i=1}^B \norm{G_i}_2\right)^2 B \norm{g - u}_2^2.
\end{align}

\section{An upper bound to the gradient norm}

In this section, we reiterate the analysis from the main paper (\S~3.2) with
more details.

Let $\theta^{(l)} \in \mathbb{R}^{M_l \times M_{l-1}}$ be the weight matrix for
layer $l$ and $\s^{(l)}(\cdot)$ be a Lipschitz continuous activation function.
Then, let
\begin{align}
    x^{(0)}         & = x  \label{eq:sup_notation_start} \\
    z^{(l)}         & = \theta^{(l)} \, x^{(l-1)}        \\
    x^{(l)}         & = \s^{(l)}(z^{(l)})                \\
    \Psi(x; \Theta) & = x^{(L)}. \label{eq:sup_notation_end}
\end{align}

Equations \ref{eq:sup_notation_start}-\ref{eq:sup_notation_end} define a simple
fully connected neural network without bias to simplify the closed form
definition of the gradient with respect to the parameters $\Theta$.

In addition we define the gradient of the loss with respect to the output of
the network as
\begin{align}
\nabla_{x_i^{(L)}} \calL & = \nabla_{x_i^{(L)}} \calL(\Psi(x_i; \Theta), y_i)
\end{align}
and the gradient of the loss with respect to the output of layer $l$ as
\begin{align}
\nabla_{x_i^{(l)}} \calL = \Delta_i^{(l)} \Si_L'(z_i^{(L)})
    \nabla_{x_i^{(L)}}\calL
\end{align}
where
\begin{align}
\Delta_i^{(l)} = \Si_l'(z_i^{(l)}) \theta_{l+1}^T \dots \Si_{L-1}'(z_i^{(L-1)}) \theta_L^T
\end{align}
propagates the gradient from the last layer (pre-activation) to layer $l$ and
\begin{align}
\Si_l'(z) = diag\left(\sigma'^{(l)}(z_1), \dots, \sigma'^{(l)}(z_{M_l})\right)
\end{align}
defines the gradient of the activation function of layer $l$.

Finally, the gradient with respect to the parameters of the $l$-th layer can be
written
\begin{eqnarray}
\lefteqn{\norm{\nabla_{\theta_l} \calL(\Psi(x_i; \Theta), y_i)}_2} \\
    & = & \norm{\left( \Delta_i^{(l)} \Si_L'(z_i^{(L)})
        \nabla_{x_i^{(L)}}\calL \right) \left(x_i^{(l-1)}\right)^T}_2 \\
    & \leq & \norm{x_i^{(l-1)}}_2 \norm{\Delta_i^{(l)}}_2
        \norm{\Si_L'(z_i^{(L)}) \nabla_{x_i^{(L)}}\calL}_2.
\end{eqnarray}
We observe that $x_i^{(l)}$ and $\Delta_i^{(l)}$ depend only on $z_i$ and
$\Theta$. However, we theorize that due to various weight initialization and
activation normalization techniques those quantities do not capture the
important per sample variations of the gradient norm. Using the above, which is
also shown experimentally to be true in \S~4.1, we deduce the following upper
bound per layer
\begin{eqnarray}
\lefteqn{\norm{\nabla_{\theta_l} \calL(\Psi(x_i; \Theta), y_i)}_2} \\
    & \leq & \max_{l, i}\left( \norm{x_i^{(l-1)}}_2
        \norm{\Delta_i^{(l)}}_2 \right) \norm{\Si_L'(z_i^{(L)})
        \nabla_{x_i^{(L)}}\calL}_2 \\
    & = & \rho \norm{\Si_L'(z_i^{(L)}) \nabla_{x_i^{(L)}}\calL}_2,
\end{eqnarray}
which can then be used to derive our final upper bound
\begin{align}
    \norm{\nabla_{\Theta} \calL(\Psi(x_i; \Theta), y_i)}_2 \leq
        \underbrace{L \rho \norm{\Si_L'(z_i^{(L)})
            \nabla_{x_i^{(L)}}\calL}_2}_{\hat{G}_i} \label{eq:sup_upper_bound}.
\end{align}

Intuitively, equation \ref{eq:sup_upper_bound} means that the variations of the
gradient norm are mostly captured by the final classification layer.
Consequently, we can use the gradient of the loss with respect to the
pre-activation outputs of our neural network as an upper bound to the
per-sample gradient norm.

\section{Comparison with SVRG methods}

For completeness, we also compare our proposed method with Stochastic Variance
Reduced Gradient methods and present the results in this section. We follow the
experimental setup of \S~4.2 and evaluate on the augmented CIFAR10 and CIFAR100
datasets. The algorithms we considered were SVRG
\cite{johnson2013accelerating}, accelerated SVRG with Katyusha momentum
\cite{allen2017katyusha} and, the most suitable for Deep Learning, SCSG
\cite{lei2017non} which in practice is a mini-batch version of SVRG. SAGA
\cite{defazio2014saga} was not considered due to the prohibitive memory
requirements for storing the per sample gradients.

\begin{figure*}
    \centering
    \begin{subfigure}[t]{0.49\textwidth}
        \includegraphics[width=\textwidth]{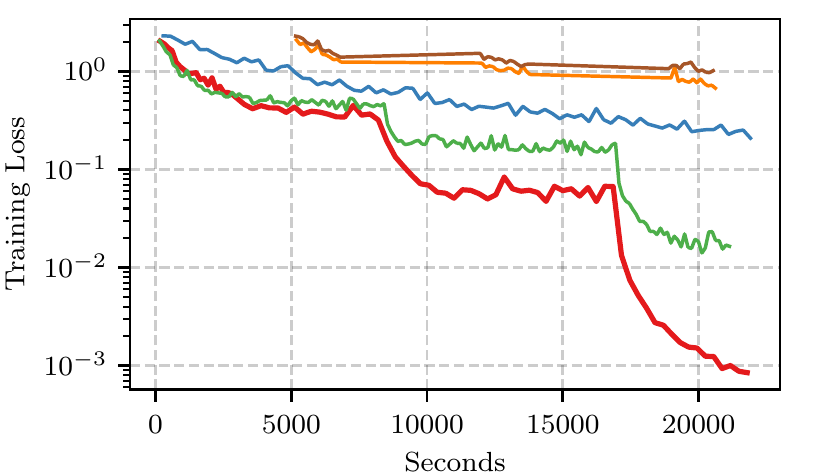}
        \caption{CIFAR10 Training Loss}
    \end{subfigure}
    ~
    \begin{subfigure}[t]{0.49\textwidth}
        \includegraphics[width=\textwidth]{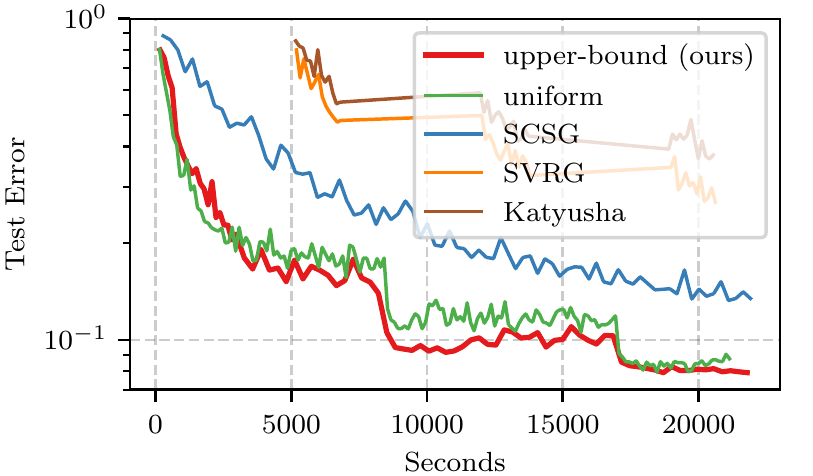}
        \caption{CIFAR10 Test Error}
    \end{subfigure} \\
    \begin{subfigure}[t]{0.49\textwidth}
        \includegraphics[width=\textwidth]{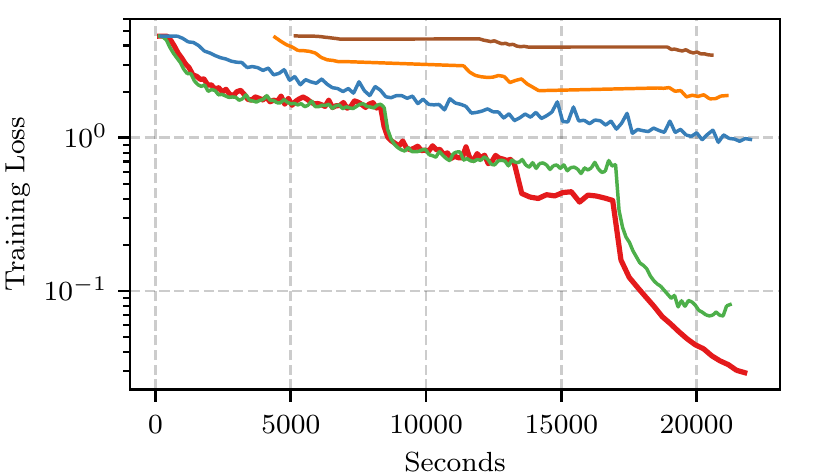}
        \caption{CIFAR100 Training Loss}
    \end{subfigure}
    ~
    \begin{subfigure}[t]{0.49\textwidth}
        \includegraphics[width=\textwidth]{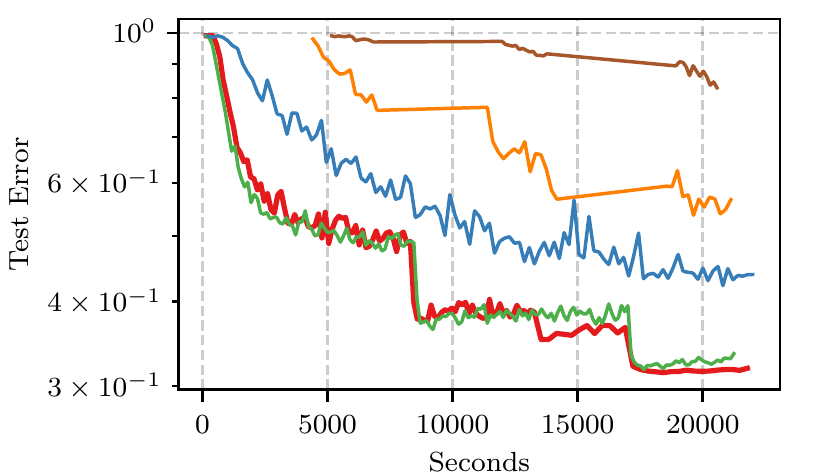}
        \caption{CIFAR100 Test Error}
    \end{subfigure}
    \caption{Comparison of our proposed importance sampling scheme
    (\emph{upper-bound}) to SGD with uniform sampling and variance reduced
    methods. Only SCSG can actually perform enough iterations to optimize the
    network. However, SGD with uniform sampling and our \emph{upper-bound}
    greatly outperform SCSG.} 
    \label{fig:svrg}
\end{figure*}

For all methods, we tune the learning rate and the epochs per batch gradient
computation ($m$ in SVRG literature). For SCSG, we also tune the large batch
(denoted as $B_j$ in \citet{lei2017non}) and its growth rate. The results are
depicted in figure~\ref{fig:svrg}. We observe that SGD with momentum performs
significantly better than all SVRG methods. Full batch SVRG and Katyusha
perform a small number of parameter updates thus failing to optimize the
networks. In all cases, the best variance reduced method achieves more than an
order of magnitude higher training loss than our proposed importance sampling
scheme.

\section{Ablation study on $B$}

The only hyperparameter that is somewhat hard to define in our algorithm is the
pre-sampling size $B$. As mentioned in the main paper, it controls the maximum
possible variance reduction and also how much wall-clock time one iteration
with importance sampling will require.

\begin{figure*}
    \begin{subfigure}{0.49\textwidth}
        \includegraphics{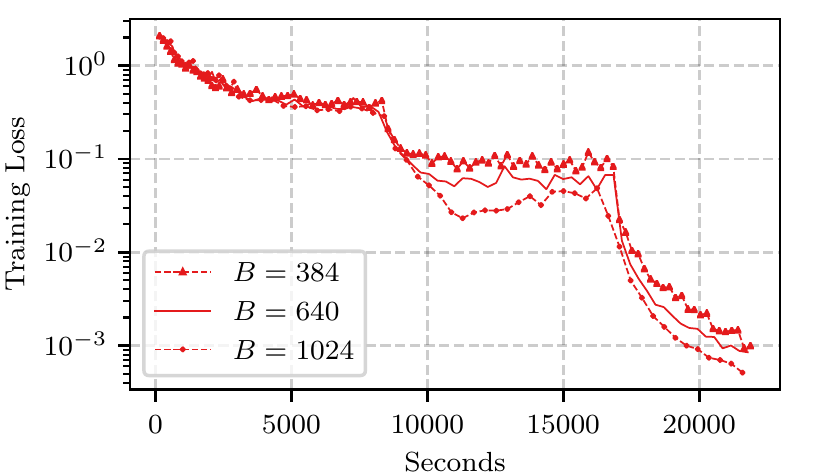}
    \end{subfigure}
    \begin{subfigure}{0.49\textwidth}
        \includegraphics{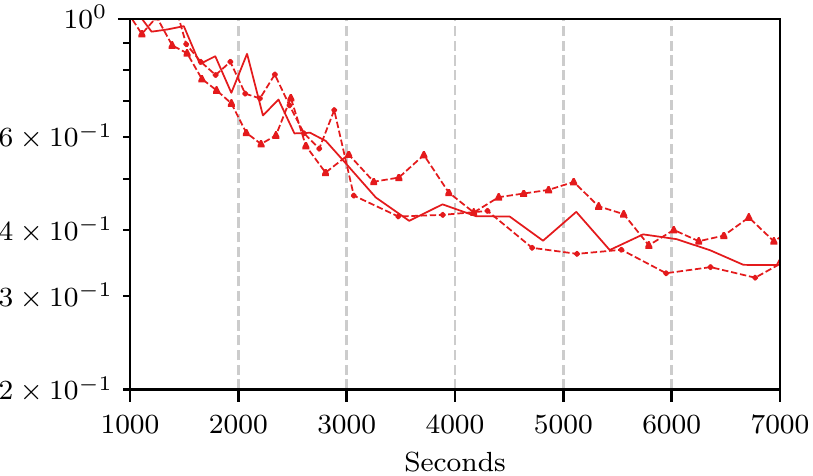}
    \end{subfigure}
    \caption{Results on training with different $B$ on CIFAR10. See the paper
    for the experimental setup.} \label{fig:b_ablation}
\end{figure*}

In figure~\ref{fig:b_ablation} we depict the results of training with
importance sampling and different pre-sampling sizes on CIFAR10. We follow the
same experimental setup as in the paper.

We observe that larger presampling size results in lower training loss, which
follows from our theory since the maximum variance reduction is smaller with
small $B$. In this experiment we use the same $\tau_{th}$ for all the methods
and we observe that $B=384$ reaches first to $0.6$ training loss. This is
justified because computing the importance for $1,024$ samples in the beginning
of training is wasteful according to our analysis.

According to this preliminary ablation study for $B$, we conclude that choosing
$B = k b$ with $2 < k < 6$ is a good strategy for achieving a speedup. However,
regardless of the choice of $B$, pairing it with a threshold $\tau_{th}$
designated by the analysis in the paper guarantees that the algorithm will be
spending time on importance sampling only when the variance can be greatly
reduced.

\section{Importance Sampling with the Loss}

In this section we will present a small analysis that provides intuition
regarding using the loss as an approximation or an upper bound to the per
sample gradient norm.

Let $\calL(\psi, y) : D \to \mathbb{R}$ be either the negative log likelihood through
a sigmoid or the squared error loss function defined respectively as
\begin{equation}
\begin{aligned}
\calL_1(\psi, y) & =  -\log\left(\frac{\exp(y\psi)}{1 + \exp(y \psi)}\right) & y \in \{-1, 1\} \quad 
    \psi \in \mathbb{R} \\
\calL_2(\psi, y) & = \norm{y - \psi}_2^2 & y \in \mathbb{R}^d \quad
    \psi \in \mathbb{R}^d
\end{aligned}
\end{equation}

Given our upper bound to the gradient norm, we can write
\begin{align}
\norm{\nabla_{\theta_t} \calL(\Psi(x_i; \theta_t), y_i)}_2 \leq L \rho
    \norm{\nabla_{\psi} \calL(\Psi(x_i; \theta_t), y_i)}_2.
    \label{eq:loss_bound}
\end{align}

Moreover, for the losses that we are considering, when $\calL(\psi, y) \to 0$
then $\norm{\nabla_{\psi} \calL(\Psi(x_i; \theta_t), y_i)}_2 \to 0$. Using this
fact in combination to equation~\ref{eq:loss_bound}, we claim that so does the
per sample gradient norm thus small loss values imply small gradients. However,
large loss values are not well correlated with the gradient norm which can also
be observed in \S~4.1 in the paper.

To summarize, we conjecture that due to the above facts, sampling
proportionally to the loss reduces the variance only when the majority of the
samples have losses close to 0. Our assumption is validated from our
experiments, where the \emph{loss} struggles to achieve a speedup in the early
stages of training where most samples still have relatively large loss values.

\end{document}